\documentclass[conference]{IEEEtran}
\IEEEoverridecommandlockouts
\usepackage{cite}
\usepackage{amsmath,amssymb,amsfonts}
\usepackage{algorithmic}
\usepackage{graphicx}
\usepackage{textcomp}
\usepackage{soul}
\usepackage{multirow}
\usepackage{booktabs}
\usepackage[table]{xcolor}
\usepackage{tabularx} 
\usepackage{tikz}
\usepackage{enumitem}

\begin{document}

\title{New Perspectives on the Use of Online Learning for Congestion Level Prediction over Traffic Data}
\author{\IEEEauthorblockN{Eric L. Manibardo, Ibai La\~na, Jesus L. Lobo}
\IEEEauthorblockA{\textit{TECNALIA, Basque Research and Technology Alliance (BRTA)} \\
48160 Derio, Bizkaia, Spain\\
\{eric.lopez, ibai.lana, jesus.lopez\}@tecnalia.com}
\and
\IEEEauthorblockN{Javier Del Ser}
\IEEEauthorblockA{\textit{University of the Basque Country (UPV/EHU)} \\
48013 Bilbao, Bizkaia, Spain \\
javier.delser@ehu.eus}
}
\maketitle

\begin{abstract}
This work focuses on classification over time series data. When a time series is generated by non-stationary phenomena, the pattern relating the series with the class to be predicted may evolve over time (concept drift). Consequently, predictive models aimed to learn this pattern may become eventually obsolete, hence failing to sustain performance levels of practical use. To overcome this model degradation, online learning methods incrementally learn from new data samples arriving over time, and accommodate eventual changes along the data stream by implementing assorted concept drift strategies. In this manuscript we elaborate on the suitability of online learning methods to predict the road congestion level based on traffic speed time series data. We draw interesting insights on the performance degradation when the forecasting horizon is increased. As opposed to what is done in most literature, we provide evidence of the importance of assessing the distribution of classes over time before designing and tuning the learning model. This previous exercise may give a hint of the predictability of the different congestion levels under target. Experimental results are discussed over real traffic speed data captured by inductive loops deployed over Seattle (USA). Several online learning methods are analyzed, from traditional incremental learning algorithms to more elaborated deep learning models. As shown by the reported results, when increasing the prediction horizon, the performance of all models degrade severely due to the distribution of classes along time, which supports our claim about the importance of analyzing this distribution prior to the design of the model.
\end{abstract}

\begin{IEEEkeywords}
Time series, online learning, deep learning, concept drift, traffic forecasting, congestion prediction.
\end{IEEEkeywords}

\section{Introduction} \label{sec:intro}

Real-time machine learning (also known as \emph{stream learning} or \emph{stream data mining}) has acquired special relevance with the advent of the Big Data era \cite{laney20013d,MOA-Book-2018}, becoming one of its most widely acknowledged challenges. Due to the incoming volume of data, their speed or the lack of computational resources, stream learning algorithms have no access to all historical stream data because the storage capacity needed for this purpose becomes unaffordable. Indeed, data streams are fast and large in size (potentially infinite), so information must be extracted from them in real time. The usage of limited resources (e.g. time and memory) often implies sacrificing performance for efficiency of the learning technique in use. 

Besides the inherent difficulty of learning from streaming data incrementally, data streams are often produced by non-stationary phenomena, which may imprint changes on the incoming data distribution, ultimately leading to the so-called \emph{concept drift} \cite{gama2014survey}. Drifts imply that predictive models trained over data become eventually obsolete, and do not adapt suitably to new distributions. Therefore, they must adapt to drifts as fast as possible to maintain good performance scores. In this context, the community has devoted intense research efforts towards the development of stream learning algorithms capable of undertaking predictive tasks over data streams under minimum time and memory requirements, and with resiliency against drifts in the stream data distribution \cite{losing2018incremental,lu2018learning,lobo2020spiking}. The need for overcoming these drawbacks stems from many real applications, such as manufacturing, environmental sensing, telecommunications, social media, marketing, entertainment, and smart grids, to mention a few \cite{vzliobaite2016overview}. 

Among such applications, one of the fields where stream learning methods have been targeted most is traffic modeling \cite{lana2018road}. Indeed, endless vehicular data are produced nowadays, coming from inductive loops hidden beneath ground soil, traffic cameras or infrared sensors. Due to its direct application in the context of traffic management, traffic forecasting by using diverse machine learning flavors conforms a very active investigation field, with a wide research community, and dozens of scientific publications every year \cite{vlahogianni2014short,nagy2018survey}. These comprehensive surveys show that, although traffic flow soars as the main traffic variable to be predicted, variables such as speed, travel time or occupation are also gaining momentum in recent years as a consequence of  their actionability as road service level estimators. The essential purpose of traffic management systems, for which the level of service is inferred from e.g. congestion levels, is to take active measures and provide information to road users \cite{horvitz2012prediction}. Thus, while traffic flow forecasts need to be interpreted alongside other inputs, such as the road capacity or the typical flow profiles at different locations of the road network, variables like speed or travel time are more straightforward to be used, as it is easier for a practitioner to discriminate whether a certain speed implies free-flow circulation or a bottleneck. 

On the other hand, traffic flow time series present recognisable daily patterns \cite{lana2016understanding} over time. Such patterns ease the formulation of short-term prediction schemes, whereas speed time series show the effects described in the three-phase traffic theory \cite{kerner2004three}, staging longer periods without change, and being changes particularly abrupt. Speed predictions are, hence, more challenging to obtain than flow estimations, although they might render better performance metrics when the error is averaged. All in all, defining congestion levels as an output variable either before or after performing the prediction is a step that helps not only with the actionability of data-informed traffic management processes, but also with the assessment of individual performance metrics for each traffic service level. 

Unfortunately, to the best of the authors' knowledge there is no prior work around the feasibility of predicting service levels from traffic data by resorting to online learning methods. In traffic management it is often the case that the legacy traffic management infrastructure does not meet the computational requirements imposed by the fast arriving data flows recorded by road sensors, nor do traditional models consider the possibility that the captured road data evolve over time. This work covers this research niche by focusing on the online estimation of traffic congestion levels by using a wide spectrum of data stream learning algorithms. Specifically, the contributions of this work can be summarized as follows:
\begin{itemize}[leftmargin=*]
    \item We design a thorough model comparison study comprising offline and online variants of well-established learning algorithms, including traditional batch learning algorithms, learning methods suited for evolving data streams, and recurrent Deep Learning approaches.
    \item We assess the performance of the aforementioned online and offline learning methods on a real speed dataset captured over Seattle (USA), comprising a variety of values for the forecasting horizon.
    \item We shed light on the specific nature of time series underneath this prediction problem, which unveil the reasons for the noted high degradation of the models under consideration when the horizon prediction is increased. 
\end{itemize}

The rest of the manuscript is structured as follows: Section \ref{sec:Materials} provides information about the data and learning methods considered in the study. Next, Section \ref{sec:Experimental Setup} presents the design of the experimental setup, whereas Section \ref{sec:Results} collects and discusses the obtained results. Finally, Section \ref{sec:conc} summarizes the conclusions drawn from this work, and outlines future research lines rooted on our findings.

\section{Materials and Methods}\label{sec:Materials}

The data utilized for experimentation have been retrieved from a public repository of vehicular traffic captured over the road network of Seattle (USA), published in \cite{cui2018deep}. The dataset provides speed measurements collected by $322$ inductive loops (also denoted as automatic traffic reader (ATR) in the specialized literature) deployed on four freeways in Seattle area: I-5, I-405, I-90, and SR-520. Readings are provided in miles per hour, every 5 minutes for the whole year 2015, and they present no missing data, amounting to $105120$ speed values for each ATR. These retrieved data are used to train methods able to predict congestion levels at different points of the road. The comparison study of predictive methods described below is tested on four locations in order to assess its performance with different traffic profiles. Two ATRs in I-5 ($153.48$ and $176.01$ mileposts), one in SR-520 ($3.97$ milepost) and last one in I-405 (7.00 milepost) have been selected, establishing as selecting principle that none of them or their surrounding ones are located immediately before or after an intersection with another freeway, which could distort relations among them. 

It is intuitive to think that congestion events are propagated downstream along the road. However, certain circumstances of traffic may have an impact upstream \cite{cassidy1999some}. Based on this rationale, we propose the following scheme: for a certain speed value of inductive loop A recorded at time step $t$, the predictive features will be assumed to be speed values at time steps $\{t-5, \dots ,t-1\}$ recorded in the $4$ next and four previous ATRs located in the road surroundings of A, as well as the past speed values $\{t-5, \dots ,t-1\}$ recorded in A itself. Thus, each data instance fed to the predictive models consists of $45$ speed input values and one output target variable. For each group of $4+4+1=9$ ATRs, the maximum distance between the center and the furthest ATR never results to be greater than $6$ miles. This allows for the analysis of spatio-temporal relations among the different ATR locations. In most recent literature, it is usual to find that Deep Learning approaches for traffic modeling take advantage of such spatio-temporal relationships \cite{yu2017spatiotemporal,zhang2017deep}. It is our hypothesis that the impact of such relations is not that straightforward. We will elaborate further on this statement with the results of our study in hands (Section \ref{sec:Results}). For the sake of space, we do not provide details for the selection of the number of ATRs before and after the target one, which was done based exclusively on their relative contribution to the performance of the models (as per an ablation study performed off-line).

Faced with the challenge of estimating congestion levels, speed data must be labeled accordingly. Under this premise, there are two possible approaches when facing a congestion level estimation problem: i) to map the time series to congestion labels under a certain criterion, yielding a discrete annotation of samples that allow undertaking the classification task directly; or ii) to formulate the estimation of the congestion level as a regression problem, predict the continuous value of the speed directly, and then apply the aforementioned criterion to the predicted value to designate its congestion label. In the first case, it is crucial to know the characteristics that define each class, or to have the time series already labeled by an external agent. Otherwise, it is only possible to perform the second method, in which the predicted continuous signal is delivered to the external agent, in order to apply the classification \textit{a posteriori}. Anyhow, there is a need to transform the speed continuous series of values into a discrete series of levels of congestion. To this end, three phases of traffic (\texttt{free-flow}, \texttt{congestion} and \texttt{bottleneck}) are established according to different thresholds of speed, following the criterion described in \cite{kerner2004three}. We have not considered any of these options better \textit{a priori}, so that experiments later discussed can shed light on which of them perform better. 

\subsection{Considered Learning Methods}\label{subsec:Bench_Methods}

Once datasets have been furnished, we design a performance comparison study comprising a selection of learning methods that are capable of operating in both online and offline (batch) mode. The applicability of these methods depend on whether the congestion level estimation strategy is approached as a classification or regression task, whose convenience in terms of performance will be analyzed with quantitative evidences in Section \ref{sec:exp_reg_clf}. We now list the considered learning methods without considering the annotation strategy, along with some customized approaches designed ad-hoc for this study:

\subsubsection{Naïve Method (NM)} 
The predicted value for the traffic congestion level of NM equals that of the last example seen by the model. 

\subsubsection{Shallow Learning methods}
Except for the last two methods (which are available in the Scikit-Learn library \cite{scikit-learn}), the bulk of shallow learning methods considered in our study are implemented in Scikit-Multiflow \cite{skmultiflow}. The Scikit-Multiflow library is designed for learning from stream data in Python, built upon other popular open-source libraries including the aforementioned Scikit-Learn \cite{scikit-learn}, MOA \cite{MOA-Book-2018} and MEKA \cite{MEKA}. It provides multiple state-of-the-art learning methods for different stream learning problems, including single-output, multi-output and multi-label predictive tasks. All shallow methods have been initialized by using its default configuration, and are next listed and described briefly:
\begin{itemize}[leftmargin=*]
    \item \emph{Naïve Bayes} (NB): a Bayesian model assuming independence between input features given the output \cite{naivebayes2017}.
    \item \emph{KNN-ADWIN} (KNNA): a K-Nearest Neighbors classifier implementing the adaptive window (ADWIN) change detector to actively adapt to drifts \cite{guo2003knn,bifet2007ADWIN}. 
    \item \emph{Perceptron} (P): a linear classifier without drift adaptation.
    \item \emph{Adaptive Random Forest} (ARF): a Random Forest with a drift detector per compounding tree, triggering selective resets in response \cite{gomes2017adaptiveRF}.
    \item \emph{Additive Expert Ensemble} (AEE): an ensemble that adapts to concept drift by adding new experts (prediction strategies), pruning the weakest ones according to a weighting policy, and predicting the output with the greatest weight \cite{kolter2005usingAEE}.
    \item \emph{Dynamic Weighted Majority} (DWM): similar to AEE, it employs different weighting policies \cite{kolter2007dynamicDWM}.
    \item \emph{Online Boosting} (OB): online version of the AdaBoost ensemble method, including ADWIN at its core \cite{wang2016onlineOB}.
    \item \emph{Online Smote Bagging} (OSB): online version of the SMOTEBagging ensemble method, including ADWIN and oversampling methods to account for class imbalance \cite{wang2016onlineOB}.
    \item \emph{Oza Bagging} (OZB): online version of the Oza Bagging ensemble method \cite{oza2005onlineOZB}.
    \item \emph{Oza Bagging-ADWIN} (OZBA): OZB variant that incorporates an ADWIN change detector.
    \item \emph{Hoeffding Tree} (HT): very fast decision tree capable of adapting to changes, that uses the Hoeffding bound to determine the number of examples needed to make a decision. It grows an alternative sub-tree whenever an old one becomes questionable, and replaces the latter when the new becomes more accurate \cite{hulten2001miningHT}.%
    \item \emph{Hoeffding Adaptive Tree} (HAT): Hoeffding bound based decision tree, using ADWIN \cite{bifet2009adaptiveHAT}.
    \item \emph{Hoeffding Anytime Tree} (HATT): extremely fast decision tree that is similar to HT, but performs new splits in the tree as soon as the improvement of making this action is proven. This makes HATT learn new concepts faster, but with greater computational load \cite{manapragada2018extremelyHATT}.
    \item \emph{Adaptive Very Fast Decision Rules} (VFDR): in contrast to nodes and leafs present in decision trees, VFDR constructs a set of rules that provides more design flexibility, as it allows for the removal of individual rules without rebuilding the entire model \cite{kosina2015veryAVFDR}.
    \item \emph{Passive Aggressive} (PA): this method splits the solution space by a weight vector. When a wrong classification takes place, the weight vector is updated (\emph{aggressive state}). Otherwise, the algorithm status does not change (\emph{passive state}) \cite{crammer2006onlinePA}.
    \item \emph{Stochastic Gradient Descent} (SGD): linear classifiers (e.g. SVM with linear kernel, logistic regression) implementing stochastic gradient descent training \cite{StochasticGradientDescentSGD}.
\end{itemize}

\subsubsection{Deep Learning methods}

Besides traditional shallow learning techniques, Deep Learning models have recently shown good capabilities for traffic forecasting \cite{chen2016long, yu2017deep,zhang2017deep, yu2017spatiotemporal}. Motivated by the conclusions drawn from these works, we have designed a Deep Learning architecture specifically designed for online learning, based on an hybridization of convolutional layers (to capture short-term time dependencies) and Long Short-Term Memory (LSTM) units (to grasp long-term dependencies over time) \cite{chen2016long, hochreiter1997long}. The aforementioned input of 45 speed features per predicted value, is initially processed through a one-dimensional convolutional layer of 32 filters and 32 time steps. The dimensionality of the output of this convolutional layer is reduced by applying MaxPooling with a factor of 2. Output values are fed to a stateful LSTM layer of 64 cells to extract long-term temporal relationships. Then, a fully-connected layer of 50 neurons connect the flattened output of the hybrid convolutional-recurrent architecture to the target variable to be predicted. Specifically, a single neuron provides the predicted speed value at time step $t+1$ for regression. In the case of classification, the last layer is composed by 3 neurons, one per feasible class, preceded by a Softmax activation function that converts logit values to estimations of the probability of each congestion level. Hyperparameters were chosen after an offline grid search. 

Along with the latter, another three architectures are considered by removing the convolutional layer, or by replacing the LSTM units of the recurrent layer with less-parametric Gated Recurrent Units (GRU) \cite{chung2014empirical}. This yields:  
\begin{itemize}[leftmargin=*]
    \item \emph{Online Convolutional LSTM Neural Network} (OCLSTM): the Deep Learning architecture described above.
    \item \emph{Online LSTM Neural Network} (OLSTM): OCLSTM without the convolutional layer.
    \item \emph{Online Convolutional GRU Neural Network} (OCGRU): OCLSTM, changing the LSTM units with GRUs.
    \item \emph{Online GRU Neural Network} (OGRU): OLSTM, changing the LSTM units with GRUs.
\end{itemize}

\subsubsection{Extreme Learning Machine} 

extreme learning machine (ELM) is a generalization of single-hidden feed-forward networks (SLFN), where the hidden layer, that carries out feature mapping, does not need to be tuned \cite{huang2011extreme}. Essentially, ELM initializes at random the values of weights and biases of hidden layer neurons, making them independent of the training data. The input data is projected into hidden layer after applying weights and biases, after which the weights between the hidden layer and the output layer can be learned efficiently by performing a generalized Moore-Penrose inversion of the matrix containing such weights. In this work we consider an online sequential implementation of ELM, known as OS-ELM \cite{liang2006fast}, with a hidden layer of 1500 units (selected after off-line fine-tuning, with results not shown due to the lack of space).

\subsection{Offline and Online Versions of the Learning Methods}

Traffic forecasting systems are commonly operated in longer intervals that those typically tackled in stream learning scenarios. This is why research done in this area is scarce \cite{niu2015online,chen2016long}. However, in the particular context of traffic forecasting, the advantages of adopting a stream learning approach reside in the need for dealing with possible concept drift \cite{lana2018road}, as well as in the implementation constraints derived from deploying the model \cite{horvitz2012prediction}. Usually, after a traditional batch training phase using all initially available data, the traffic forecasting system is deployed. Then new streaming data arrives, but it can be unfeasible to retrain and update the model by simply learning again from scratch over all data received until then. 

When this is the case, two options can be selected: i) keeping the original model, without updating it whatsoever (offline); or ii) incrementally learning from every newly arriving sample (online). To this end, we compare different offline/online learning methods, from traditional learning algorithms that allow for incremental learning (including those designed for concept drift adaptation), to more elaborated novel deep learning architectures. This will allow us to compare between both approaches, while analyzing the advantages and caveats of each learning mode for the problem at hand. In addition, the naïve model is used as the baseline of our comparison study, which sets the minimum hurdle the other learning methods should overpass to justify their adoption.

\subsection{Classification Metrics}\label{sec:metrics}

The accuracy of the estimated congestion levels is evaluated in terms of the $F_1$ score \cite{goutte2005probabilistic}. A separated $F_1^l$ score is computed for each of the three congestion levels $l\in\{\texttt{free-flow},\texttt{congestion},\texttt{bottleneck}\}$. Since the annotation of the dataset gives rise to a severely imbalanced distribution of classes (congestion levels), we opt for an unweighted mean $UMF_1$ of the aforementioned $F_1^l$ scores:
\begin{equation}
 UMF_1=\frac{1}{3}\sum_{l}{F_1}^l, \label{eq1}
\end{equation}
where all classes feature equal importance (weight) in the computation. Consequently, the value of the overall score does not get affected by a skewed distribution of the score across different classes.

\section{Experimental Setup}\label{sec:Experimental Setup}

This section describes the experimental setup constructed to provide an informed answer to three different research questions (RQ):
\begin{itemize}[leftmargin=*]
    \item RQ1: Should we tackle the problem as a classification task, or instead predict the speed value and discretize afterwards?
    \item RQ2: Which online learning technique performs best?
    \item RQ3: How does the performance degrade when the forecasting horizon is increased? Why? 
\end{itemize}

\begin{figure*}[ht]
\includegraphics[width=1.9\columnwidth]{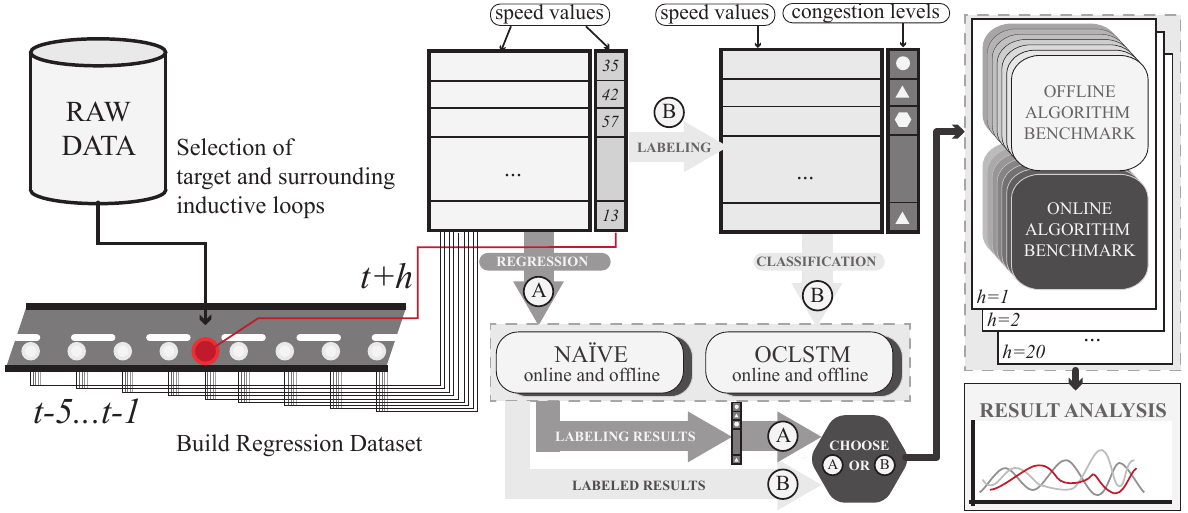}
\centering
\caption{Experimentation workflow. Datasets are built based on speed values, from selected mileposts of a highway. Then, we use the na\"ive model NM and the OCLSTM model to choose among one of two strategies: predict speed values and then apply labeling process \protect\textcircled{\raisebox{-0.5pt}A} or to label the original dataset prior to modeling and then, to face a classification problem \protect\textcircled{\raisebox{-0.8pt}B}. After selecting one of these strategies, a comparison study of different learning methods and forecasting horizons is performed for offline/online modes of operation.}
\label{fig:process}
\end{figure*}

Before attempting to answer RQ2 and RQ3, it is compulsory to select, as per the response to RQ1, one of the main paths of the workflow displayed in Figure \ref{fig:process}. To this end, in order to discriminate between regression (\textcircled{\raisebox{-0.5pt}A}) or classification (\textcircled{\raisebox{-0.8pt}B}), we use OCLSTM architecture, in conjunction with NM as a baseline of the worst case scenario. These two methods should achieve different levels of predictive performance among regression and classification, which will allow us to discard one of these strategies, and proceed forward with the rest of the study by focusing only on the strategy of choice.

After selecting between \textcircled{\raisebox{-0.5pt}A} or \textcircled{\raisebox{-0.8pt}B} strategies, the rest of the experiments are executed with the overall best performing strategy. In regards to RQ2, for each learning method under consideration, we compare its performance when operating in offline (no update) and online fashion (incremental learning). Initially, the algorithms are trained with only the first week of year 2015, employing a batch size of one sample in order to be able to update the model after each arriving sample, during online setting. Because recordings are detached by 5 minutes, this yields 2016 examples for the batch training phase. For the online setting, the model predicts the next arriving sample as a test, and assumes that the real value of the example just tested is available for training (\emph{test-then-train}). This scheme is held until the last sample of the dataset is reached. The offline setting proceeds in the same way, but without updating the model after every new sample. By comparing these approaches, we can reveal the degree of improvement between both options.

In addition to the above offline/online comparison study, each model is tested with different prediction horizons (RQ3), from $h=1$ (i.e. prediction for slot $t$ with features up to time $t-1$) to $h=20$, implying the estimation of the speed at the evaluated point within $5$ minutes and up to $100$ minutes in the future. A deeper horizon would imply adopting different forecasting strategies, focused on the long term, such as clustering or historical averaging, instead of the pattern-based models here considered. This evaluation procedure gives rise to a set of performance values from $h=1$ to $h=20$, for predictions obtained with speed measurements obtained in instants $\{t-5, \dots ,t-1\}$ from the surrounding ATRs, and from the ATR under analysis itself. The analysis of different predictive horizons is crucial for this experimental setup, as the way in which predictions  degrade can be indicative of how the input features relates to the output to be estimated.

\section{Results and Discussion}\label{sec:Results}

Prior to applying the experimental workflow defined in Section \ref{sec:Experimental Setup}, thresholds that define the three established classes must be determined. Following the guidelines described in \cite{kerner2004three} and after analysing in detail the traffic profiles of the roads under study, thresholds have been established in 42 mph, as the speed over which the state is considered free flow, and 22 mph, as the speed under which the speed is considered a bottleneck. Congested state is defined between both thresholds. When these cutting lines are plotted over data, one can annotate sampling points with their traffic congestion level at every point in time: as such, samples labeled with \texttt{free-flow} and \texttt{bottleneck} are above and below the defined levels, while those belonging to the \texttt{congestion} class fall in between.

\subsection*{RQ1: Should we tackle the Problem as a Classification Task, or instead predict the Speed Value and discretize afterwards?}\label{sec:exp_reg_clf}

Experiments to answer RQ1 are run in the first place. Table \ref{tab:table1} reports the obtained results for ATR SR520, from which we extract conclusions buttressed by the scores achieved for the rest of ATRs (not shown for the sake of clarity). 
\begin{table}[h!]
\caption{Performance of regression \protect\textcircled{\raisebox{-0.5pt}A} and classification \protect\textcircled{\raisebox{-0.8pt}B} strategies, using offline and online settings ($F_1$-score per class and overall $UMF_1$)}
\label{tab:table1}
\resizebox{\columnwidth}{!}{
\begin{tabular}{rcccccccc} \toprule
  \multirow{2}{*}{\textbf{\textit{Methods}}} &
 \multicolumn{2}{c}{\texttt{Freeflow}}  &
 \multicolumn{2}{c}{\texttt{Congestion}}   &
 \multicolumn{2}{c}{\texttt{Bottleneck}}  &
 \multicolumn{2}{c}{Overall}   \\
  \cmidrule(lr){2-3}\cmidrule(lr){4-5}\cmidrule(lr){6-7}\cmidrule(lr){8-9}
   & \textit{Offline}   & \textit{Online}    & \textit{Offline}   & \textit{Online}    & \textit{Offline}   & \textit{Online}    & \textit{Offline}   & \textit{Online} \\
\cmidrule(lr){1-9}
NM & 0.986  & 0.986 & 0.738   & 0.738 & 0.614   & 0.614    & 0.779   & 0.779  \\
 \multirow{1}{*}{\textcircled{\raisebox{-0.5pt}A}  }
 OCLSTM & 0.946 & 0.985 & 0 & 0.746 & 0 & 0.482 & 0.315 & 0.738\\
 \multirow{1}{*}{\textcircled{\raisebox{-0.8pt}B}  }
 OCLSTM & 0.981  & 0.992 & 0.387   & 0.845 & 0.417   & 0.755    & 0.595   & 0.864  \\
 \bottomrule
 \end{tabular}
}
\end{table}

The NM exhibit performance scores considered to be the baseline of any other modeling choice. Any more complex learning method whose efficiency is underneath this performance is of no practical value, as NM does not require any training, nor does it demand computational efforts to produce a prediction. Being NM a na\"ive model without any learning capabilities, offline and online performance scores are the same in this case. When inspecting the OCLSTM results, we should bear in mind that in highways, the dominant class is typically \texttt{free-flow}. Our collected data is not an exception to this statement. For this reason, even at an offline setting, $F_1$-score is high for this class (traffic is mostly \texttt{free-flow}), leaving slight room for improvement during online learning. However, the other two classes are more scarce, so they become a key point when attempting to boost performance. 

Table \ref{tab:table1} reveals that the discretization of speed values prior to modeling (classification strategy) improves the accuracy of the estimations for both offline and online settings. This suggests that given only 2016 examples during initial batch training phase, regression is a more difficult task with respect to classification, leading to worse performance for \textcircled{\raisebox{-0.5pt}A} if no model update is done. It could be expected that, given enough data and updates to the model during online training, the score achieved by the regression and classification strategies would eventually converge. However, in a realistic setting, strategy \textcircled{\raisebox{-0.8pt}B} (classification) was found to surpass \textcircled{\raisebox{-0.5pt}A} (regression) over all classes. This is the reason why \textcircled{\raisebox{-0.8pt}B} is hereafter adopted as the best strategy for the application under consideration.

\subsection*{RQ2: Which Learning Technique performs best?}
\begin{table*}[ht!]
\centering
\caption{\label{tab:table2}$UMF_1$ of different learning techniques over $t+1$ forecasting horizon. Column titles show analyzed interval.}
\resizebox{1.5\columnwidth}{!}{
\begin{tabular}{ccccccccc} 
 \toprule
  \multirow{2}{*}{\textbf{\textit{Methods}}} 
  &
 \multicolumn{2}{c}{\textbf{I-405 \scriptsize{(4.73-8.90 mile)}}}  &
 \multicolumn{2}{c}{\textbf{I-5 \scriptsize{(151.25-155.69 mile)}}}   &
 \multicolumn{2}{c}{\textbf{I-5 \scriptsize{(174.16-177.75 mile)}}}   &
 \multicolumn{2}{c}{\textbf{SR-520 \scriptsize{(0.83-5.14 mile)}}} \\
 \cmidrule(lr){2-3}\cmidrule(lr){4-5}\cmidrule(lr){6-7}\cmidrule(lr){8-9}
 
 & \textit{Offline}   & \textit{Online}    & \textit{Offline}   & \textit{Online}    & \textit{Offline}   & \textit{Online}    & \textit{Offline}   & \textit{Online} \\
 \cmidrule(lr){1-9}
 NB & 0.722  & 0.736  & 0.677   & 0.755  & 0.683   & 0.704   & 0.630   & 0.760 \\
KNNA &  0.729   & 0.738    & 0.742    & 0.766 & 0.729   & 0.727   & 0.585  & 0.766  \\
P & 0.615   & 0.818   & 0.410  & 0.729 & 0.424   & 0.682   & 0.379  & 0.709    \\
ARF &  0.805 & 0.961  & 0.748   & 0.960 & 0.639   & 0.945   & 0.610   & 0.948   \\
AEE & 0.736  & 0.739  & 0.683  & 0.759 & 0.707  & 0.701    & 0.653 & 0.763  \\
DWM &  0.725 & 0.743 & 0.686   & 0.772 & 0.683   & 0.714    & 0.653   & 0.751   \\
OB & 0.271 & 0.801  & 0.319   & 0.838 & 0.321   & 0.807    & 0.315   & 0.800    \\
OSB &  0.271 & 0.687 & 0.319  & 0.722 & 0.321   & 0.670    & 0.315   & 0.765   \\
OZB & 0.722  & 0.737   & 0.747   & 0.763 & 0.728   & 0.727    & 0.594   & 0.765    \\
OZBA &  0.271  & 0.730 & 0.319   & 0.763 & 0.321   & 0.718   & 0.315   & 0.762   \\
VFDR & 0.819 & 0.993   & 0.677   & 0.980 & 0.683   & 0.971    & 0.630   & 0.978    \\
HT &  0.892 & 0.995 & 0.677   & 0.980 & 0.683   & 0.984    & 0.630   & 0.994   \\
HAT & 0.572   & 0.788  & 0.668   & 0.851 & 0.676   & 0.739   & 0.634   & 0.856    \\
HATT &  0.807  & 0.992 & 0.319   & 0.984 & 0.804   & 0.985 & 0.592   & 0.991   \\
PA & 0.660    & 0.751   & 0.412  & 0.679 & 0.507   & 0.621   & 0.565 & 0.723    \\
SGD &  0.437   & 0.832 & 0.490   & 0.737 & 0.353   & 0.691    & 0.479   & 0.794   \\
OELM & 0.507   & 0.835 & 0.650   & 0.857 & 0.452   & 0.631    & 0.564   & 0.840    \\
OCLSTM & 0.829  & 0.890 & 0.716   & 0.805 & 0.647   & 0.848    & 0.595   & 0.864    \\
OLSTM &  0.833  & 0.914   & 0.737   & 0.826 & 0.743   & 0.899   & 0.583   & 0.724   \\
OCGRU & 0.809   & 0.873  & 0.764   & 0.794 & 0.600   & 0.829   & 0.738   & 0.831    \\
OGRU &  0.841  & 0.910   & 0.678  & 0.745 & 0.756   & 0.885    & 0.576   & 0.700   \\
NM & 0.728   & 0.728  & 0.772 & 0.772 & 0.694   & 0.694     & 0.779   & 0.779    \\
 \bottomrule
\end{tabular}
}
\end{table*}
After selecting classification \textcircled{\raisebox{-0.8pt}B} as the preferred strategy for traffic congestion estimation, we have performed several simulations over the data gathered in the four mileposts defined in Section \ref{sec:Materials}. For each point, UMF\textsubscript{1} between the three congestion levels is computed, applying offline and online settings, over all learning methods described at Subsection \ref{subsec:Bench_Methods}. The results are reported in Table \ref{tab:table2}, which depicts a test bench of different learning methods and the degree of improvement that would be expected when adopting and online approach over offline.

Again, NM shares the same metrics for the offline/online settings, and sets a lower performance bound to be over-passed by the rest of models. With one week of data provided at offline setting, and by never updating these models, it is \textit{a priori} expected that online versions perform better, with continuously updated knowledge. However, we observe that these performance differences are not that acute for most methods, which entails that the one week information that the offline models lean on is enough to provide reliable predictions. Differences in which this gap operates in each location reveals how stable data are: for instance, differences are slighter in I-405 than in SR-520, which suggests that speed data behaves more consistently along weeks in the former ATR than in the latter. This poses a question about how much information is required to build a reliable offline model at each location.

When online approaches are considered, it should be noted that ARF, HT, HATT and VFDR present the highest performance, showing that adaptative learning methods represent the best approach when dealing with online stream data problems. The first three methods operate quite similar, by growing a new and more adapted-to-actual-trend decision tree and replacing older one. VFDR uses rules instead of decision trees, but concept drift adaptation mechanism works quite similar to each other. In the case of OELM based solution, it performs well at three of the analyzed ATR locations, but at the last one it falls below NM metrics. OELM is a learning method that adapts quickly to changes, with minimal computational cost, but at a high dependence on the hidden layer initialization, which usually delivers unstable results. Lastly, the designed Deep Learning architectures perform beyond NM in the online setting. However, we would like to note that there is not a best architecture for this specific problem. At some cases, models featuring the convolutional layer (i.e. OCLSTM, OCGRU) render higher scores, while at other cases, plain recurrent neural networks occur to perform better. Consequently, in no way we can claim that LSTM based Deep Learning methods outperform those relying over GRU-based counterparts or vice-versa, because there is no pattern that support any of these hypotheses. 

\subsection*{RQ3: How does the performance degrade when the forecasting horizon is increased? Why?}\label{sec:horizon_results}

\begin{figure*}[h]
\includegraphics[width=1.8\columnwidth]{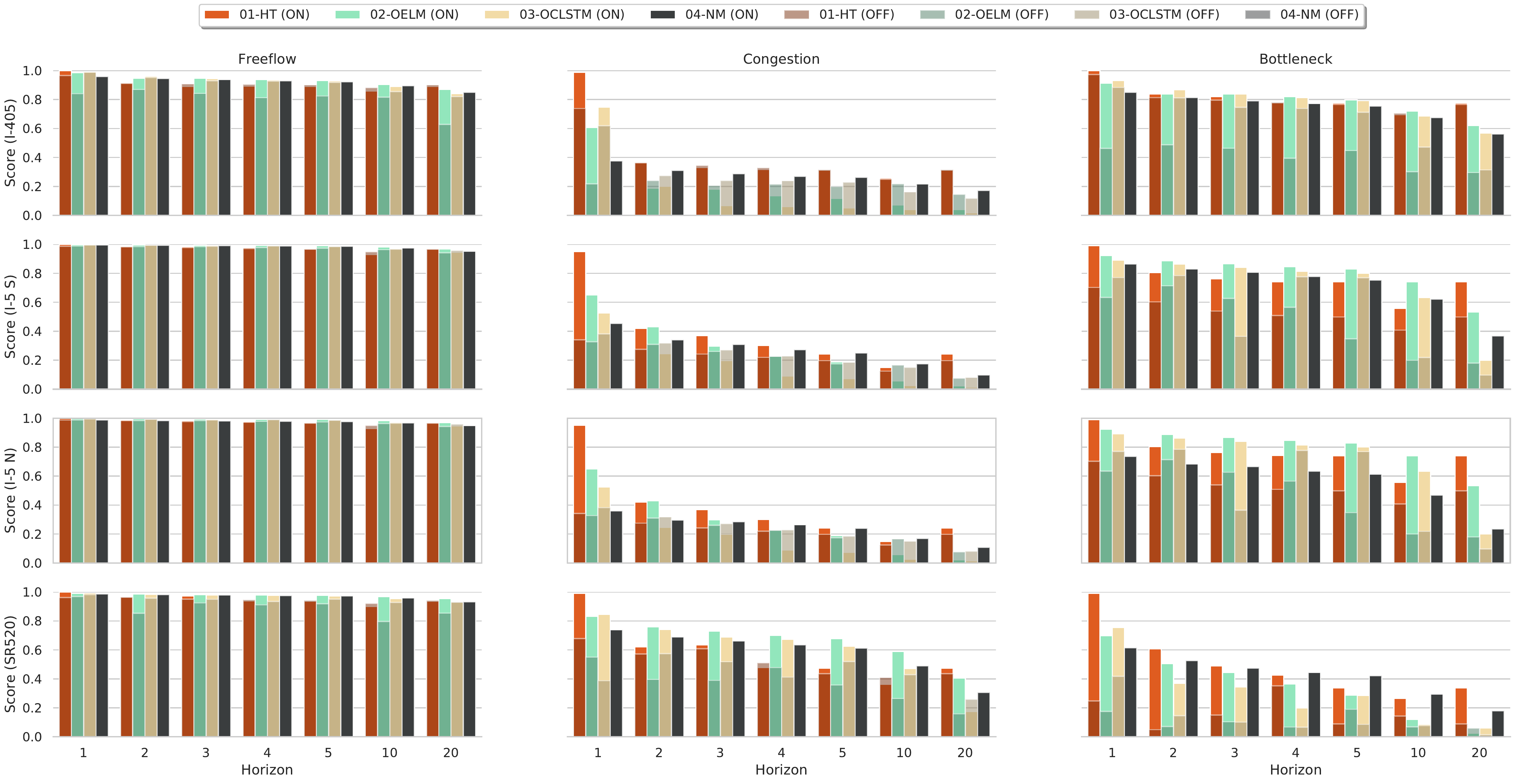}
\centering
\caption{Performance per class (columns) and ATR locations (rows) for the selected learning methods, over offline and online settings. The upper legend indicates chromatic identifier of each model implementation, placing offline on top of online settings, due to their lower performance. Seven increasing forecasting horizons are analyzed, with the interval between horizons being set to 5 minutes.}
\label{fig:Horizon}
\end{figure*}

Finally, we also analyzed the performance evolution of the models when the forecasting horizon is increased. For this purpose, we have selected the best performing method of each family presented in Section \ref{subsec:Bench_Methods}. Then, we test offline/online settings for $h=1$, $h=5$, $h=10$ and $h=20$. Figure \ref{fig:Horizon} collects graphically the obtained results.

As could have been expected beforehand, at three of the four considered mileposts, the performance score of the \texttt{congestion} class decreases drastically after the forecasting horizon is set beyond $h=1$. Given the described class thresholds, while analyzing speed data series, it is common that there are no more than two consecutive congestion labeled samples, before class changes from \texttt{free-flow} to \texttt{bottleneck} or from \texttt{bottleneck} to \texttt{free-flow}. The instability of class distribution makes the prediction of the \texttt{congestion} class particularly challenging for higher forecasting horizons, due to a lack of available information for models: when models learn from examples of this class, most of their recent past correspond to speed values that belong to other class. In the test phase, a sample with those kind of values is more likely to be classified in the other class, specially having the model much more samples of this kind to observe. However, those transitions (from \texttt{free-flow}/\texttt{bottleneck} to \texttt{congestion}) represent the real challenge of estimating speed, as from a predictive perspective, estimating the next data point is for most of data series as simple as providing the previous data point: in \texttt{free-flow}, the speed will be mostly the free-flow speed, and during a bottleneck, it will be close to 0. Thus, speed estimations that are obtained with regression techniques and assessed by measuring the error could be regarded as impractical, and certainly pointless. Indeed, in many cases the good performance metrics respond to the abundance of free-flow periods in the series, in which the error is minimal, while for the periods of change, bigger errors are produced, but they are dissipated when all error measurements are averaged. This effect can be seen in our own experiments when errors per class are averaged into UMF$_1$. The interest, hence, resides in the detection of the moment when a transition between states is produced.

In the case of the SR520 highway, traffic profile is clearly different with respect the other analyzed mileposts. Here, the performance of \texttt{bottleneck} class has a more pronounced negative trend when increasing the forecasting horizon. Our hypothesis is that it is a less crowded highway where the worst congestion level is rarely reached for a long period of time. As most of the observed samples correspond to \texttt{free-flow} and \texttt{congestion}, it is harder for the model to predict correctly the \texttt{bottleneck} state. Nevertheless, as Figure \ref{fig:Horizon} clearly shows, it is easier for a model to face a lack of consecutive examples of the \texttt{bottleneck} class when compared to \texttt{congestion}, as this latter class occurs in short transitions between the other traffic levels. In other words, the traffic state needs to pass through the \texttt{congestion} class in order to switch between \texttt{free-flow} and \texttt{bottleneck}, making mistakes when predicting the \texttt{congestion} class the most harmful for the overall performance of the model.

\section{Conclusions and Future Work}\label{sec:conc}

In this paper we have revealed the capabilities of online learning setting for congestion level prediction over traffic data. The provided results support the initial hypothesis of the remarkable degree of improvement that online learning should achieve over offline learning, when traditional batch training approach is not an option. Furthermore, additional analysis when enlarging the forecasting horizon was performed by considering the best performing approach as per a comparison study of diverse offline/online learning methods. These outcomes support our initial speculations about the importance of undertaking a previous study on the class distribution, before starting to develop a classification model. Unfortunately, many scientific results reported in the literature are exposed with an inappropriate approach, where scores of the majority class disguise the poor prediction performance of other classes. However, the real practical value of these models is to excel at identifying transitions between classes, because it is in this shift when the road traffic profile changes. 

All these insights open up a line of future work that could exploit the observed similarities between traffic profiles at different points of the road network. Specifically, transfer learning could help in this regard, by which models specific to a certain highway point are not pre-trained in batch, but rather take advantage of other models already trained elsewhere over the road network. This approach could reduce the amount of training data required to properly develop a predictive model.

\section*{Acknowledgments}

The authors would like to thank the Basque Government for its funding support through the EMAITEK and ELKARTEK programs. Eric L. Manibardo receives funding support from the Basque Government through its BIKAINTEK PhD support program (grant no. 48AFW22019-00002). 

\bibliographystyle{./bibliography/IEEEtran}
\bibliography{./bibliography/IEEEexample}

\begin{thebibliography}{10}
\providecommand{\url}[1]{#1}
\csname url@samestyle\endcsname
\providecommand{\newblock}{\relax}
\providecommand{\bibinfo}[2]{#2}
\providecommand{\BIBentrySTDinterwordspacing}{\spaceskip=0pt\relax}
\providecommand{\BIBentryALTinterwordstretchfactor}{4}
\providecommand{\BIBentryALTinterwordspacing}{\spaceskip=\fontdimen2\font plus
\BIBentryALTinterwordstretchfactor\fontdimen3\font minus
  \fontdimen4\font\relax}
\providecommand{\BIBforeignlanguage}[2]{{%
\expandafter\ifx\csname l@#1\endcsname\relax
\typeout{** WARNING: IEEEtran.bst: No hyphenation pattern has been}%
\typeout{** loaded for the language `#1'. Using the pattern for}%
\typeout{** the default language instead.}%
\else
\language=\csname l@#1\endcsname
\fi
#2}}
\providecommand{\BIBdecl}{\relax}
\BIBdecl

\bibitem{laney20013d}
D.~Laney, ``{3D} data management: Controlling data volume, velocity and
  variety,'' \emph{META group research note}, vol.~6, no.~70, p.~1, 2001.

\bibitem{MOA-Book-2018}
A.~Bifet, R.~Gavald\`a, G.~Holmes, and B.~Pfahringer, \emph{Machine Learning
  for Data Streams with Practical Examples in MOA}.\hskip 1em plus 0.5em minus
  0.4em\relax MIT Press, 2018.

\bibitem{gama2014survey}
J.~Gama, I.~{\v{Z}}liobait{\.e}, A.~Bifet, M.~Pechenizkiy, and A.~Bouchachia,
  ``A survey on concept drift adaptation,'' \emph{ACM computing surveys
  (CSUR)}, vol.~46, no.~4, p.~44, 2014.

\bibitem{losing2018incremental}
V.~Losing, B.~Hammer, and H.~Wersing, ``Incremental on-line learning: A review
  and comparison of state of the art algorithms,'' \emph{Neurocomputing}, vol.
  275, pp. 1261--1274, 2018.

\bibitem{lu2018learning}
J.~Lu, A.~Liu, F.~Dong, F.~Gu, J.~Gama, and G.~Zhang, ``Learning under concept
  drift: A review,'' \emph{IEEE Transactions on Knowledge and Data
  Engineering}, 2018.

\bibitem{lobo2020spiking}
J.~L. Lobo, J.~Del~Ser, A.~Bifet, and N.~Kasabov, ``Spiking neural networks and
  online learning: An overview and perspectives,'' \emph{Neural Networks}, vol.
  121, pp. 88--100, 2020.

\bibitem{vzliobaite2016overview}
I.~{\v{Z}}liobait{\.e}, M.~Pechenizkiy, and J.~Gama, ``An overview of concept
  drift applications,'' in \emph{Big Data Analysis: New Algorithms for a New
  Society}.\hskip 1em plus 0.5em minus 0.4em\relax Springer, 2016, pp. 91--114.

\bibitem{lana2018road}
I.~La\~na, J.~Del~Ser, M.~Velez, and E.~I. Vlahogianni, ``Road traffic
  forecasting: recent advances and new challenges,'' \emph{IEEE Intelligent
  Transportation Systems Magazine}, vol.~10, no.~2, pp. 93--109, 2018.

\bibitem{vlahogianni2014short}
E.~I. Vlahogianni, M.~G. Karlaftis, and J.~C. Golias, ``Short-term traffic
  forecasting: Where we are and where we’re going,'' \emph{Transportation
  Research Part C: Emerging Technologies}, vol.~43, pp. 3--19, 2014.

\bibitem{nagy2018survey}
A.~M. Nagy and V.~Simon, ``Survey on traffic prediction in smart cities,''
  \emph{Pervasive and Mobile Computing}, vol.~50, pp. 148--163, 2018.

\bibitem{horvitz2012prediction}
E.~J. Horvitz, J.~Apacible, R.~Sarin, and L.~Liao, ``Prediction, expectation,
  and surprise: Methods, designs, and study of a deployed traffic forecasting
  service,'' \emph{arXiv preprint arXiv:1207.1352}, 2012.

\bibitem{lana2016understanding}
I.~La\~na, J.~Del~Ser, and I.~n. Olabarrieta, ``Understanding daily mobility
  patterns in urban road networks using traffic flow analytics,'' in
  \emph{IEEE/IFIP Network Operations and Management Symposium}, 2016, pp.
  1157--1162.

\bibitem{kerner2004three}
B.~S. Kerner, ``Three-phase traffic theory and highway capacity,''
  \emph{Physica A: Statistical Mechanics and its Applications}, vol. 333, pp.
  379--440, 2004.

\bibitem{cui2018deep}
Z.~Cui, R.~Ke, and Y.~Wang, ``Deep bidirectional and unidirectional lstm
  recurrent neural network for network-wide traffic speed prediction,''
  \emph{arXiv preprint arXiv:1801.02143}, 2018.

\bibitem{cassidy1999some}
M.~J. Cassidy and R.~L. Bertini, ``Some traffic features at freeway
  bottlenecks,'' \emph{Transportation Research Part B: Methodological},
  vol.~33, no.~1, pp. 25--42, 1999.

\bibitem{yu2017spatiotemporal}
H.~Yu, Z.~Wu, S.~Wang, Y.~Wang, and X.~Ma, ``Spatiotemporal recurrent
  convolutional networks for traffic prediction in transportation networks,''
  \emph{Sensors}, vol.~17, no.~7, p. 1501, 2017.

\bibitem{zhang2017deep}
J.~Zhang, Y.~Zheng, and D.~Qi, ``Deep spatio-temporal residual networks for
  citywide crowd flows prediction,'' in \emph{AAAI Conference on Artificial
  Intelligence}, 2017.

\bibitem{scikit-learn}
F.~Pedregosa, G.~Varoquaux, A.~Gramfort, V.~Michel, B.~Thirion, O.~Grisel,
  M.~Blondel, P.~Prettenhofer, R.~Weiss, V.~Dubourg, J.~Vanderplas, A.~Passos,
  D.~Cournapeau, M.~Brucher, M.~Perrot, and E.~Duchesnay, ``Scikit-learn:
  Machine learning in {P}ython,'' \emph{Journal of Machine Learning Research},
  vol.~12, pp. 2825--2830, 2011.

\bibitem{skmultiflow}
J.~Montiel, J.~Read, A.~Bifet, and T.~Abdessalem, ``Scikit-multiflow: A
  multi-output streaming framework,'' \emph{Journal of Machine Learning
  Research}, vol.~19, no.~72, pp. 1--5, 2018.

\bibitem{MEKA}
J.~Read, P.~Reutemann, B.~Pfahringer, and G.~Holmes, ``{MEKA}: A
  multi-label/multi-target extension to {Weka},'' \emph{Journal of Machine
  Learning Research}, vol.~17, no.~21, pp. 1--5, 2016.

\bibitem{naivebayes2017}
P.~Kaviani and S.~Dhotre, ``Short survey on naive bayes algorithm,''
  \emph{International Journal of Advance Research in Computer Science and
  Management}, vol.~04, 11 2017.

\bibitem{guo2003knn}
G.~Guo, H.~Wang, D.~Bell, Y.~Bi, and K.~Greer, ``Knn model-based approach in
  classification,'' in \emph{OTM Confederated International Conferences ``On
  the Move to Meaningful Internet Systems''}.\hskip 1em plus 0.5em minus
  0.4em\relax Springer, 2003, pp. 986--996.

\bibitem{bifet2007ADWIN}
A.~Bifet and R.~Gavalda, ``Learning from time-changing data with adaptive
  windowing,'' in \emph{SIAM International Conference on Data Mining}, 2007,
  pp. 443--448.

\bibitem{gomes2017adaptiveRF}
H.~M. Gomes, A.~Bifet, J.~Read, J.~P. Barddal, F.~Enembreck, B.~Pfharinger,
  G.~Holmes, and T.~Abdessalem, ``Adaptive random forests for evolving data
  stream classification,'' \emph{Machine Learning}, vol. 106, no. 9-10, pp.
  1469--1495, 2017.

\bibitem{kolter2005usingAEE}
J.~Z. Kolter and M.~A. Maloof, ``Using additive expert ensembles to cope with
  concept drift,'' in \emph{International Conference on Machine Learning},
  2005, pp. 449--456.

\bibitem{kolter2007dynamicDWM}
------, ``Dynamic weighted majority: An ensemble method for drifting
  concepts,'' \emph{Journal of Machine Learning Research}, vol.~8, no. Dec, pp.
  2755--2790, 2007.

\bibitem{wang2016onlineOB}
B.~Wang and J.~Pineau, ``Online bagging and boosting for imbalanced data
  streams,'' \emph{IEEE Transactions on Knowledge and Data Engineering},
  vol.~28, no.~12, pp. 3353--3366, 2016.

\bibitem{oza2005onlineOZB}
N.~C. Oza, ``Online bagging and boosting,'' in \emph{IEEE International
  Conference on Systems, Man and Cybernetics}, vol.~3, 2005, pp. 2340--2345.

\bibitem{hulten2001miningHT}
G.~Hulten, L.~Spencer, and P.~Domingos, ``Mining time-changing data streams,''
  in \emph{ACM SIGKDD International Conference on Knowledge Discovery and Data
  Mining}, 2001, pp. 97--106.

\bibitem{bifet2009adaptiveHAT}
A.~Bifet and R.~Gavald{\`a}, ``Adaptive learning from evolving data streams,''
  in \emph{International Symposium on Intelligent Data Analysis}.\hskip 1em
  plus 0.5em minus 0.4em\relax Springer, 2009, pp. 249--260.

\bibitem{manapragada2018extremelyHATT}
C.~Manapragada, G.~I. Webb, and M.~Salehi, ``Extremely fast decision tree,'' in
  \emph{ACM SIGKDD International Conference on Knowledge Discovery \& Data
  Mining}, 2018, pp. 1953--1962.

\bibitem{kosina2015veryAVFDR}
P.~Kosina and J.~Gama, ``Very fast decision rules for classification in data
  streams,'' \emph{Data Mining and Knowledge Discovery}, vol.~29, no.~1, pp.
  168--202, 2015.

\bibitem{crammer2006onlinePA}
K.~Crammer, O.~Dekel, J.~Keshet, S.~Shalev-Shwartz, and Y.~Singer, ``Online
  passive-aggressive algorithms,'' \emph{Journal of Machine Learning Research},
  vol.~7, no. Mar, pp. 551--585, 2006.

\bibitem{StochasticGradientDescentSGD}
L.~Bottou, ``Large-scale machine learning with stochastic gradient descent,''
  in \emph{Proceedings of COMPSTAT'2010}, Y.~Lechevallier and G.~Saporta,
  Eds.\hskip 1em plus 0.5em minus 0.4em\relax Heidelberg: Physica-Verlag HD,
  2010, pp. 177--186.

\bibitem{chen2016long}
Y.~Chen, Y.~Lv, Z.~Li, and F.-Y. Wang, ``Long short-term memory model for
  traffic congestion prediction with online open data,'' in \emph{IEEE
  International Conference on Intelligent Transportation Systems}, 2016, pp.
  132--137.

\bibitem{yu2017deep}
R.~Yu, Y.~Li, C.~Shahabi, U.~Demiryurek, and Y.~Liu, ``Deep learning: A generic
  approach for extreme condition traffic forecasting,'' in \emph{SIAM
  International Conference on Data Mining}, 2017, pp. 777--785.

\bibitem{hochreiter1997long}
S.~Hochreiter and J.~Schmidhuber, ``Long short-term memory,'' \emph{Neural
  computation}, vol.~9, no.~8, pp. 1735--1780, 1997.

\bibitem{chung2014empirical}
J.~Chung, C.~Gulcehre, K.~Cho, and Y.~Bengio, ``Empirical evaluation of gated
  recurrent neural networks on sequence modeling,'' \emph{arXiv preprint
  arXiv:1412.3555}, 2014.

\bibitem{huang2011extreme}
G.-B. Huang, H.~Zhou, X.~Ding, and R.~Zhang, ``Extreme learning machine for
  regression and multiclass classification,'' \emph{IEEE Transactions on
  Systems, Man, and Cybernetics, Part B (Cybernetics)}, vol.~42, no.~2, pp.
  513--529, 2011.

\bibitem{liang2006fast}
N.-Y. Liang, G.-B. Huang, P.~Saratchandran, and N.~Sundararajan, ``A fast and
  accurate online sequential learning algorithm for feedforward networks,''
  \emph{IEEE Transactions on Neural Networks}, vol.~17, no.~6, pp. 1411--1423,
  2006.

\bibitem{niu2015online}
X.~Niu, Y.~Zhu, Q.~Cao, X.~Zhang, W.~Xie, and K.~Zheng, ``An
  online-traffic-prediction based route finding mechanism for smart city,''
  \emph{International Journal of Distributed Sensor Networks}, vol.~11, no.~8,
  p. 970256, 2015.

\bibitem{goutte2005probabilistic}
C.~Goutte and E.~Gaussier, ``A probabilistic interpretation of precision,
  recall and f-score, with implication for evaluation,'' in \emph{European
  Conference on Information Retrieval}, 2005, pp. 345--359.

\end{thebibliography}
\vfill
\end{document}